\documentclass[10pt,twocolumn,letterpaper]{article}

\usepackage{iccv}              % To produce the CAMERA-READY version
\usepackage{cuted}
\usepackage{graphicx}
\usepackage{colortbl}
\usepackage{multirow}
\usepackage{booktabs}
\usepackage{xcolor}
\definecolor{iccvblue}{rgb}{0.21,0.49,0.74}
\usepackage[pagebackref,breaklinks,colorlinks,allcolors=iccvblue]{hyperref}
\usepackage{siunitx}

\def\paperID{9572} % *** Enter the Paper ID here
\def\confName{ICCV}
\def\confYear{2025}

\title{Degradation-Modeled Multipath Diffusion for Tunable Metalens Photography}

\author{Jianing Zhang\\
\and
Jiayi Zhu\\
\and
Feiyu Ji\\
\and
Xiaokang Yang\\
\and
Xiaoyun Yuan\\
 }

\begin{document}

\maketitle
\begin{strip}
\begin{minipage}{\textwidth}
\vspace{-40pt}
\centering
\includegraphics[width=0.95\textwidth]{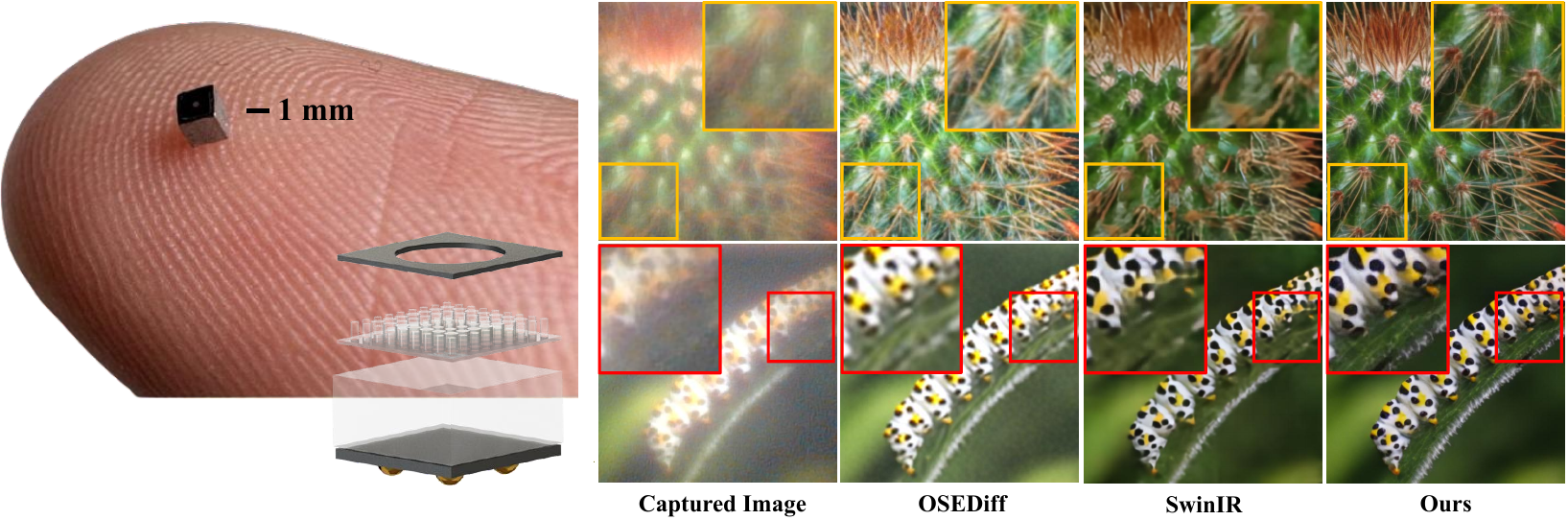}
% \captionof{figure}{Diffusion-Powered Nanooptics: A metasurface-based ultra-compact camera system empowered by a large-model-driven image restoration algorithm. Compared to state-of-the-art image restoration methods, our approach achieves superior reconstruction quality and enhanced visual perception.}
\captionof{figure}{A metasurface-based ultra-compact camera system empowered by a large-model-driven image restoration algorithm. Compared to state-of-the-art image restoration methods, our approach achieves superior reconstruction quality and enhanced visual perception.}
\label{fig:teaser}
\vspace{-5pt}
\end{minipage}
\end{strip}
\begin{abstract}

Metalenses offer significant potential for ultra-compact computational imaging but face challenges from complex optical degradation and computational restoration difficulties.
Existing methods typically rely on precise optical calibration or massive paired datasets, which are non-trivial for real-world imaging systems. Furthermore, a lack of control over the inference process often results in undesirable hallucinated artifacts.
We introduce Degradation-Modeled Multipath Diffusion for tunable metalens photography, leveraging powerful natural image priors from pretrained models instead of large datasets. Our framework uses positive, neutral, and negative-prompt paths to balance high-frequency detail generation, structural fidelity, and suppression of metalens-specific degradation, alongside \textit{pseudo} data augmentation. A tunable decoder enables controlled trade-offs between fidelity and perceptual quality. Additionally, a spatially varying degradation-aware attention (SVDA) module adaptively models complex optical and sensor-induced degradation.
Finally, we design and build a millimeter-scale MetaCamera for real-world validation. Extensive results show that our approach outperforms state-of-the-art methods, achieving high-fidelity and sharp image reconstruction. More materials: \textit{https://dmdiff.github.io/}.

% Metalenses hold great promise for ultra-compact computational imaging systems but are challenged by complex optical degradation and the difficulties of computational restoration.
% %
% Existing methods typically depend on precise optical calibration or large amounts of paired training data, non-trivial for real imaging systems. Additionally, the lack of control over the inference process often leads to undesirable hallucinated artifacts.
% %
% In this work, we introduce Degradation-Modeled Multipath Diffusion for tunable metalens photography, obtaining the powerful nature image priors from pretrained models instead of data. Our multipath diffusion framework leverages positive, neutral, and negative-prompt paths to effectively balance high-frequency detail generation, structural fidelity preservation, and suppression of metalens-style degradation alongside \textit{pseudo} training data augmentation. This is combined with a tunable decoder that enables controllable trade-offs between fidelity and perceptual quality. A spatially varying degradation-aware attention (SVDA) module is proposed to adaptively model the complex optical and sensor-induced degradation. 
% %
% Finally, we design and construct a millimeter-scale MetaCamera prototype for real-world validation. Extensive results demonstrate that our approach reconstructs images with high fidelity and sharpness, surpassing current state-of-the-art methods.

\end{abstract}    

\section{Introduction}
The demand for compact, high-performance imaging systems has grown rapidly with advancements in biomedical imaging, implantable vision systems, point-of-care diagnostics, and augmented/virtual reality (AR/VR) \cite{li2022ultracompact, li2021meta, stoja2021improving, yoon2021printable}.
However, conventional compound lens systems rely on multi-element optics to correct aberrations, making it challenging to achieve both miniaturization and high imaging quality.
Metalenses, a class of engineered optical interfaces composed of subwavelength-scale nanostructures, enable precise optical wavefront control by manipulating phase, amplitude, and polarization within an ultrathin form factor. Combined with computational restoration methods, their unique capabilities offer new opportunities for developing compact, high-performance imaging systems. \cite{gu2023reconfigurable, kim2021dielectric, li2022ultracompact, li2021meta, stoja2021improving, yoon2021printable}.

While metalenses present a promising solution for ultra-compact imaging systems, their practical implementation is constrained by optical limitations and computational complexities. Specifically, images captured by metalenses suffer from spatially varying degradations due to aberrations. Conventional model-based restoration techniques, such as Wiener filtering \cite{wiener1964extrapolation} and iterative deconvolution methods \cite{cho2009fast, krishnan2009fast, xu2013unnatural}, typically assume uniform distortions and struggle to correct the highly localized aberrations caused by the metalens. Although dividing images into small patches can mitigate this issue, it leads to reduced robustness and stability in the algorithms. %Furthermore, accurately calibrating the optical aberrations of an imaging system is a non-trivial, time-consuming, and labor-intensive task.
Deep learning-based approaches \cite{aieta2015multiwavelength, lin2014dielectric, wang2018broadband, yu2014flat} have demonstrated more significant potential for restoration, offering powerful capabilities. However, these approaches typically require large datasets of pixel-aligned image pairs for training, which are often difficult to obtain for computational imaging systems. As a result, recent research has shifted toward employing large-scale models pretrained on extensive datasets as priors for image super-resolution and restoration tasks, leading to significant improvements in robustness, generalization, and overall performance.
Among the various techniques, generative diffusion models (DMs) have demonstrated exceptional performance \cite{ho2020denoising, saharia2022image, li2022srdiff, antipa2017diffusercam, wu2024seesr}. These models, pretrained on vast collections of image and text data, possess powerful natural image priors that allow them to synthesize vivid and authentic details more effectively than conventional GAN-based methods \cite{goodfellow2014generative}. However, despite their impressive capabilities, DMs are not devoid of limitations. Their restoration process is inherently prone to generating hallucinated details that do not exist in the original image, due to the lack of fine-grained control over the restoration process during inference. 

In this manuscript, we propose a degradation-modeled multipath diffusion model for tunable metalens photography, leveraging powerful natural image priors from pre-trained models instead of large datasets. 
We first introduce the spatially varying degradation-aware attention (SVDA) module, which quantifies degradation by analyzing both the optical point spread function (PSF) of the metalens and the quality of image patches. This allows it to effectively account for degradation from both the metalens and image sensors. The degradation characterization results are further employed by the attention network to guide the LoRA fine-tuning process, enabling the diffusion model to adaptively handle region-specific degradations.
We then propose the multipath diffusion model comprising three distinct pathways: a positive-prompt path for generating high-frequency details, a neutral-prompt path for restoring structural fidelity, and a negative-prompt path for mitigating metalens-specific degradation. The negative-prompt path also facilitates the generation of \textit{pseudo} input-output paired samples, which expand the training dataset and enhance generalization. 
To enable smooth control over the reconstruction results, we introduce a tunable decoder that adaptively merges the latent codes from the positive and neutral paths, allowing users to adjust the balance between reconstruction fidelity and perceptual quality.
% We then design the multipath diffusion model consisting of a positive-prompt path to generate high-frequency details, a neutral-prompt path to restore high-fidelity structural information, and a negative-prompt path to learn and avoid the metalens-style degradation. Besides, the negative-prompt path can also generate \textit{pseudo} input-output paired samples to expand the training dataset and improve generalization ability. Correspondingly, an instantly tunable decoder is proposed to adaptively merge the latent codes of the positive and neutral paths. It provides a user-adjustable feature to smoothly control the reconstruction results, allowing for a dynamic balance between objective reconstruction fidelity and subjective perceptual quality.
%
% Next, we introduce the spatially varying degradation-aware attention (SVDA) module. Specifically, SVDA quantifies degradation by analyzing both the optical point spread function (PSF) of the metalens and the quality assessment of image patches, allowing it to effectively account for degradation from both the metalens and image sensors.
%
Finally, as demonstrated in Fig.~\ref{fig:teaser}, we design and build a micro metalens-based camera (MetaCamera) to validate our algorithm. MetaCamera achieves a size of approximately $1$ mm$^3$. The experimental results demonstrate that our algorithm successfully reconstructs images with both high fidelity and sharpness, outperforming state-of-the-art approaches.
Collectively, the contributions of this manuscript are as follows:
\begin{itemize}
    \item \textbf{Multipath Diffusion Model.} We propose a multipath diffusion model incorporating positive, neutral, and negative-prompt paths to balance high-frequency details and structural fidelity while mitigating metalens-specific degradation. Additionally, we introduce an instantly tunable decoder that enables a smooth trade-off between perceptual quality and reconstruction fidelity.
    \item \textbf{Spatially Varying Degradation Aware Attention (SVDA) Module.} We introduce the SVDA module, which quantifies degradation from both metalens aberrations and image sensors, guiding the LoRA process to fine-tune the diffusion network for adaptive handling of spatially varying degradations.
    \item \textbf{Hardware verification.} 
    We design and build a millimeter-scale metalens-based camera (MetaCamera) to verify our algorithm, demonstrating superior image reconstruction results with high fidelity and sharpness, outperforming state-of-the-art methods.
\end{itemize}

\section{Related work}
\label{sec:intro}

%-------------------------------------------------------------------------
\subsection{Flat Computational Cameras}

Researchers have explored various approaches to reduce the height and complexity of conventional compound camera optical systems. One such approach simplifies the typical multi-element optical stack into a single refractive element \cite{heide2013high, li2021universal, schuler2013machine, tanida2001thin}, effectively minimizing both geometric and chromatic aberrations. 
Another approach uses a thin monolithic sensor array \cite{venkataraman2013picam} with a color-filtered single-lens element array to mitigate chromatic aberrations, converting the deconvolution problem into a color light field reconstruction challenge.
However, solving this challenge without introducing artifacts remains difficult, which limits the potential for thin camera designs. Lensless cameras \cite{antipa2017diffusercam, khan2020flatnet, monakhova2020spectral, white2020silicon} replace the entire optical stack with an amplitude mask or diffuser, perturbing the incident wavefront. While this allows for thin cameras just a few millimeters thick, it complicates the recovery of high-quality images. 
% To overcome these challenges, Praneeth et al. \cite{chakravarthula2023thin} proposed a controllable metasurface lens array, which enables a shorter back focal length without introducing globally supported aberrations or reducing light efficiency. 

\subsection{Metasurface Optics}

Recent advancements in nanofabrication techniques have enabled the development of ultrathin metasurfaces composed of subwavelength scatterers \cite{engelberg2020advantages, froch2024beating, lin2014dielectric, mait2020potential}. 
Each scatterer within the metasurface can be independently engineered to precisely control the amplitude, phase, and polarization of incident wavefronts, a capability that has significantly advanced research in planar optical imaging \cite{chakravarthula2023thin, chen2022planar, froch2024beating}. 
For instance, Tseng et al. \cite{tseng2021neural} proposed a differentiable design framework to achieve full-color imaging with a large aperture of 0.5 mm. 
Moreover, Chakravarthula et al. \cite{chakravarthula2023thin} introduced a metalens array design that enhances image quality across the full broadband spectrum over a 100° FoV without increasing the back focal length, offering significant potential for miniaturizing wafer-level multi-element compound lens cameras. 
Recently, Lee et al. \cite{lee2024metaformer} developed a computational framework to correct aberrations in metalens-captured images, offering a promising solution to overcome the limitations of current meta-optical imaging systems.

\subsection{Image Restoration}
Image restoration aims to reconstruct high-quality images from degraded inputs. Traditional techniques like Wiener deconvolution \cite{wiener1964extrapolation} tackle this by solving inverse problems but struggle with noise and complex degradations, limiting real-world use. Deep learning has revolutionized the field, with models like SRCNN \cite{dong2014learning} and SwinIR \cite{liang2021swinir} achieving state-of-the-art results in tasks such as denoising \cite{chen2021pre, fan2022sunet, li2023efficient, tu2024ipt}, deblurring \cite{kong2023efficient, liu2024deblurdinat, wang2022uformer}, and super-resolution \cite{chen2023activating, chen2023dual, lu2022transformer}. However, a major drawback of existing methods, including SwinIR, is their dependence on simplified degradation models, which often fail to reflect real-world complexities. Therefore, recent studies have increasingly focused on generative models for image restoration, which can be broadly categorized into model-driven and prior-driven methods. Model-driven methods, such as SR3 \cite{saharia2022image} and SRDiff \cite{li2022srdiff}, adapt diffusion models to tasks like super-resolution by directly learning the mapping from degraded to high-quality images. In contrast, prior-driven methods leverage pre-trained models as priors to enhance the performance of image restoration tasks. For instance, DiffBIR \cite{antipa2017diffusercam} and SeeSR \cite{wu2024seesr} can significantly reduce computational overhead while maintaining high-quality restoration results. More recently, OSEDiff \cite{wu2025one} and S3Diff\cite{zhang2024degradation} introduced efficient one-step diffusion networks using trainable LoRA layers. Despite these promising advancements, a significant limitation of these methods is their reliance on well-defined degradation models, which restricts their applicability to computational imaging tasks under realistic and uncontrolled conditions. 
% Addressing this limitation remains a critical challenge for future research in the field. 
\section{Method}

\begin{figure*}[ht]
    \centering
    \includegraphics[width=1.0\textwidth]{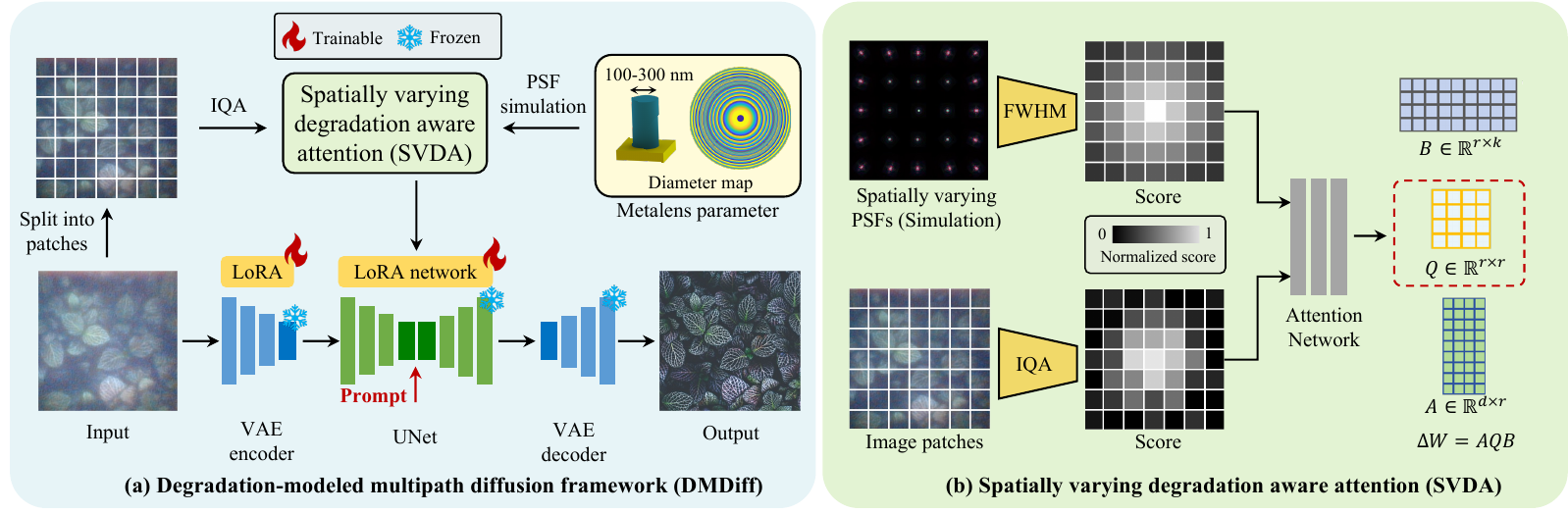} % 设置图片宽度为页面宽度
    \caption{Degradation-modeled multipath diffusion framework (DMDiff). (a) The network architecture of DMDiff. The core of DMDiff is a diffusion-based large model, where LoRA fine-tuning is applied to the encoder and UNet. (b) To address the spatially varying degradation, a spatially varying degradation aware attention (SVDA) module is proposed to guide the LoRA fine-tuning process. }
    \label{fig:multipath_diffusion_model}
    \vspace{-10pt}
\end{figure*}

To tackle the challenges of metalens-based imaging, we propose a degradation-modeled multipath diffusion framework that leverages pretrained large-scale generative diffusion models for tunable metalens photography. Our approach addresses three key challenges: complex metalens degradations, limited paired training data, and hallucinations in generative models.

With the powerful natural image priors from the base generative diffusion model, our method reconstructs vivid and realistic images using a small training dataset. To further enhance restoration, we propose a Spatially Varying Degradation-Aware (SVDA) attention module, which quantifies optical aberrations and sensor-induced noise to guide the restoration process. Additionally, we introduce a Degradation-modeled Multipath Diffusion (DMDiff) framework, incorporating positive, neutral, and negative-prompt paths to balance detail enhancement and structural fidelity while mitigating metalens-specific distortions. Finally, we design an instantly tunable decoder, enabling dynamic control over reconstruction quality to suppress hallucinations.

We begin in Section 3.1 with a description of our DMDiff with LoRA-based fine-tuning, which serves as the foundation of our approach. In Section 3.2, we introduce the SVDA module, explaining how it characterizes and leverages spatially varying degradations for adaptive restoration. Finally, in Section 3.3, we detail the multi-prompt training and inference of DMDiff, as well as the instantly tunable decoder.

% In this section, we present a framework for metalens image restoration using a pre-trained stable diffusion model. To address spatially varying degradation from fabrication errors and optical aberrations in metasurface-based imaging systems, we introduce Spatially Varying Degradation Modeled Attention to guide LoRA finetuning. To enhance imaging robustness, we propose Multipath Diffusion Training, which enables the network to restore low-frequency information and high-resolution details while learning the degradation model to expand the dataset. Finally, we introduce a tunable decoding technique based on neural coded images that allows adjustable diffusion intensity to balance subjective perception and objective accuracy during imaging.

\subsection{Degradation-modeled multipath diffusion framework}

% Recent approaches incorporating ControlNet and LoRA fine-tuning leverage generative priors from diffusion models like Stable Diffusion to improve restoration. Building on this, we propose a single-step enhancement framework with LoRA fine-tuning.

The network architecture DMDiff is presented in Fig.~\ref{fig:multipath_diffusion_model}, consisting of a VAE encoder, a latent diffusion UNet, a VAE decoder, and our spatially varying degradation aware attention module.
Here, we employ SD-Turbo \cite{sauer2024adversarial}, a distilled version of Stable Diffusion \cite{rombach2021highresolution}, as the base encoder, UNet and decoder. 
Given an input image \(I\) captured by our MetaCamera, the model encodes \(I\) into the latent space via the VAE encoder \(E\), producing the latent coded image \(z\).
Next, the UNet iteratively denoises the latent coded image \(z\) for \(k\) times, leveraging the given text prompt and spatially varying degradation cues from our SVDA module. This step ensures that the denoised latent coded image \(z\) aligns with natural image priors while mitigating metalens-specific degradation.
the VAE decoder \(D\) converts the refined latent coded image \(z\) back into a high-quality reconstructed image. In this manuscript, we set \(k=1\) to realize efficient one-step reconstruction.

To adapt the powerful pre-trained model for metasurface photography, we incorporate trainable LoRA \cite{hu2022lora} layers for fine-tuning the VAE encoder and latent diffusion UNet. %The fine-tuned networks are denoted as \(\tilde{E}\) and \(\tilde{\Phi}\), respectively. 
LoRA is an efficient fine-tuning method that updates model parameters by adding low-rank adaptation matrices to specific layers of a pretrained model. In Transformer-based architectures, LoRA decomposes the original weight matrix \( W \in \mathbb{R}^{d \times k} \) into two trainable low-rank matrices \( A \in \mathbb{R}^{d \times r} \) and \( B \in \mathbb{R}^{r \times k} \), where \( r \ll \min(d, k) \), yielding the modified weight representation:
\begin{equation}
W^* = W + AB.
\end{equation}
During training, only \( A \) and \( B \) are updated, while \( W \) remains frozen. This approach drastically reduces the number of trainable parameters, making the fine-tuning process more computationally efficient. At inference, the transformed weight \( W^* \) is integrated into the model without adding computational overhead.

\subsection{Spatially Varying Degradation Aware Attention}

\begin{figure*}[ht]
    \centering
    \includegraphics[width=0.9\textwidth]{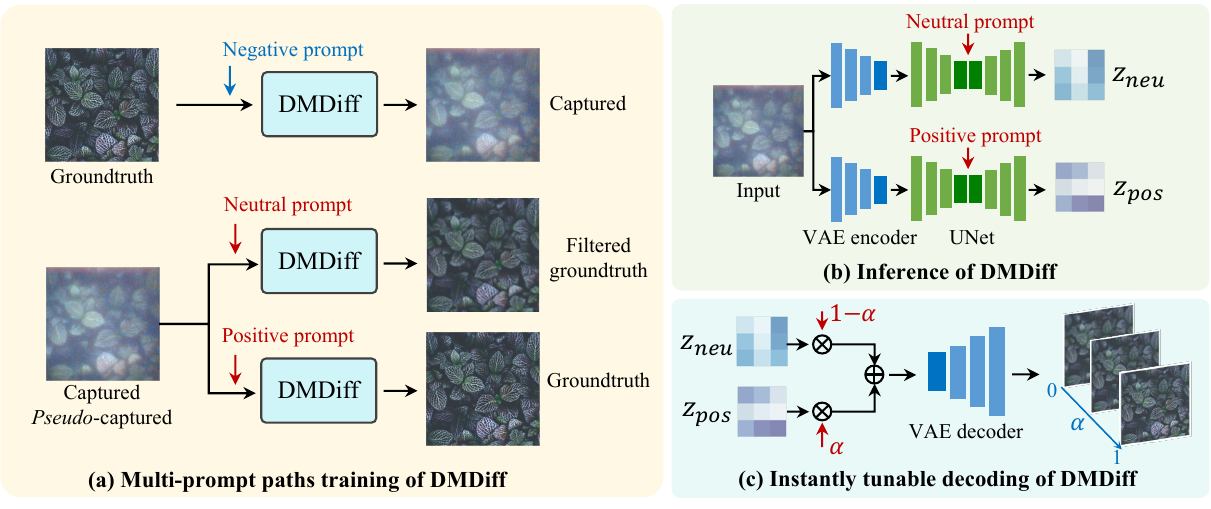} % 设置图片宽度为页面宽度
    \caption{(a) Multi-prompt paths training algorithm for DMDiff, including three paths with positive, neutral, and negative prompts. (b) In the inference step of DMDiff, the latent coded images of the neutral and positive paths are obtained. (c) Instantly tunable decoding of DMDiff, images with varying diffusion intensities can be generated quickly from the latent coded images. }
    \label{fig:training_and_inference}
    \vspace{-12pt}
\end{figure*}

Image degradation in metalens-based imaging systems primarily arises from two sources: optical aberrations and sensor-induced electronic noise, especially in miniaturized sensors. Conventional correction methods rely on physics-based models, which require precise aberration characterization. While optical aberrations can be simulated, fabrication errors and unknown illumination often lead to discrepancies between simulated and real-world distortions. Calibrating the point spread function (PSF) is also tedious and inaccurate, as it depends heavily on illumination (see Supplementary Material for calibration results).
To overcome these limitations, we propose the Spatially Varying Degradation Aware (SVDA) attention module, which enables the network to learn degradation characteristics and correct them adaptively. Instead of depending on an accurate PSF, SVDA quantifies degradation by analyzing optical degradation from simulated PSFs and incorporating quality assessments (QA) from metalens-captured images. Meanwhile, degradation introduced by the electronic sensor is also quantified using QA results. This degradation representation is then incorporated into the LoRA fine-tuning process via an attention mechanism, allowing the diffusion model to focus on region-specific degradations and enhance restoration accordingly. By leveraging SVDA, our method eliminates the need for precise aberration calibration while enabling more robust and data-driven image reconstruction.

% By fine-tuning Stable Diffusion with the proposed framework, we achieve high-resolution metasurface image restoration. While leveraging priors from large models, simple fine-tuning alone is insufficient to address non-uniform aberrations and limited training data. Aberrations from metasurface design and fabrication degrade imaging quality. As shown in Fig.~X, repeated patterns cause spatially varying degradation, with edge regions suffering the most.
% To address this, we first quantify spatial non-uniformity. A common approach estimates the Modulation Transfer Function (MTF) from the Point Spread Function (PSF), but it ignores noise and fails under severe degradation. Chromatic aberrations and sensor noise further limit SSIM and PSNR. 

To quantify optical aberrations, we use the Full Width at Half Maximum (FWHM) as a metric to assess the degradation. The FWHM is defined as the width of a distribution, specifically the distance between the points where the function reaches half of its maximum value.
The image is first divided into \( n \times n \) patches, assuming the point spread function (PSF) is constant within each patch. The PSF at the center of each patch is then simulated using Fourier optics based on the designed metalens parameters, and its FWHM is calculated by fitting a 2D Gaussian distribution to the corresponding PSF image:
\begin{equation}
  \begin{split}
      \min_{\mu_x, \mu_y, \sigma_x, \sigma_y} \| I_{\text{PSF}} - G(\mu_x, \mu_y, \sigma_x, \sigma_y) \|_2^2, \\
    \text{FWHM} = \sqrt{(2.35\sigma_x)^2 + (2.35\sigma_y)^2},
  \end{split}
\end{equation}
where \(I_{\text{PSF}}\) denotes the PSF image, and \(G(\cdot)\) represents a 2D Gaussian distribution with mean \((\mu_x, \mu_y)\) and standard deviations \((\sigma_x, \sigma_y)\). Given the radial symmetry of our metalens, the PSF image can be rotated to align its principal axes with the x- and y-directions before fitting. The FWHM of a Gaussian distribution function is \(2\sqrt{2 \ln 2} \sigma\ \approx 2.35 \sigma\).

% where \(I_{PSF}\) indicates the PSF image, \(G(\cdot)\) dentoes the 2D Gaussian distribution function with mean value \({\mu_x, \mu_y}\) and standard deviation \({\sigma_x, \sigma_y}\). As our metalens is radially symmetric, the PSF image can be rotated to align its principal axes with the x- and y-directions. 

% along the two principal axes of the fitted Gaussian function.

% \begin{equation}
%   \min_{\mu_1, \mu_2, \sigma_1, \sigma_2} \sum \| I_{PSF} - G(\mu_1, \mu_2, \sigma_1, \sigma_2) \|_2^2,
% \end{equation}

% \begin{equation}
%   \text{FWHM} = \sqrt{ \left( 2\sqrt{2\ln 2} \, \sigma_1 \right)^2 + \left( 2\sqrt{2\ln 2} \, \sigma_2 \right)^2 } ,
% \end{equation}

% \xiaoyun{
% For a given PSF image of a patch, the local luminance is defined using intensity values \( I(x, y) \) within a circular region centered at \( (x_0, y_0) \) with radius \( r \):
% \begin{equation}
% I_r = \frac{1}{N_r} \sum_{(x,y) \in C_r} I(x,y),
% \end{equation}
% where \( C_r \) represents the pixels within the circle, and \( N_r \) is their count. Varying \( r \) provides an intensity profile.
% %
% The optical imaging quality of a given image patch can be quantified using the FWHM, which is defined as:
% \begin{equation}
% \text{FWHM} = r_2 - r_1,
% \end{equation}
% where \( r_1 \) and \( r_2 \) are the smallest and largest radii satisfying \( I_r = I_{\max}/2 \). Normalizing the FWHM distribution removes intensity bias, effectively capturing spatial variations induced by optical aberrations.
% }

% While FWHM quantifies blur, it overlooks contrast inconsistencies, noise, and frequency imbalances. 

To quantify the remaining degradation, we introduce No-Reference Image Quality Assessment (NR-IQA) using the transformer-based MUSIQ metric\cite{ke2021musiq} to measure the degradation directly from the captured image patches.
Similarly, the image is divided into \( n \times n \) patches, and MUSIQ generates an image quality score distribution to capture spatial variations. Combining FWHM with NR-IQA enables a comprehensive assessment of spatially varying degradation.
%
% A key aspect of fine-tuning is incorporating degradation awareness into the LoRA framework, enabling the model to adapt to spatial variations in image quality. 
Finally, to incorporate degradation awareness into fine-tuning, we introduce an attention mechanism that utilizes FWHM and NR-IQA scores. The adaptation LoRA process is formulated as:
\begin{equation}
    \begin{aligned}
        Q &= \mathcal{N}_A(S_f, S_i) \\
        W^* &= W + A Q B
    \end{aligned}
\end{equation}
where the Attention Network (\( \mathcal{N}_A \)) takes the FWHM score (\( S_f \)) and NR-IQA score (\( S_i \)) as input and generates an \( r \times r \) attention matrix \( Q \). This mechanism allows the fine-tuning process to be guided by spatially varying degradation characteristics. By dynamically adjusting adaptation weights, the model better preserves high-quality details while correcting degraded regions, improving robustness against spatial quality variations in our MetaCamera.

\subsection{Multipath Diffusion Training}

The large-scale generative diffusion model provides powerful natural image priors for detail enhancement but also carries a high risk of undesirable hallucinated artifacts. To address this, we propose a multi-prompt training and inference algorithm for DMDiff that balances detail enhancement with fidelity preservation while mitigating metalens-specific degradation.

Specifically, our training algorithm consists of three paths with positive, neutral, and negative prompts, as illustrated in Fig.~\ref{fig:training_and_inference}(a). The neutral and positive paths take the MetaCamera-captured image as input, with the positive path learning to reconstruct the high-quality groundtruth image and the neutral path learning to reconstruct a filtered groundtruth image. Here, an edge-preserving low-pass filter is applied to the groundtruth image to remove high-frequency details while preserving strong structural content.  
In contrast, the negative path takes the high-quality groundtruth image as input and learns to generate a degraded version that mimics a MetaCamera-captured image. The three paths are conditioned on different text prompts, which are provided in the supplementary material.  

The neutral-prompt path enables dynamic control between subjective perceptual quality and objective accuracy (detailed in Section 3.4).  
The negative-prompt path helps the network learn and avoid metalens-style degradation. Additionally, since capturing pixel-aligned training pairs with a real MetaCamera is time-consuming, this reverse diffusion process can generate \textit{pseudo} input-ground truth pairs, effectively expanding the dataset and improving generalization. The effectiveness of this path is validated in the ablation study.
The training process follows a unified framework, expressed as:  
\begin{equation}
\begin{cases} 
\text{DMDiff}_{\text{neg}}(I_{gt}, P_{\text{neg}}) \rightarrow I_{in}, & \text{if } M = 1, \\
\text{DMDiff}_{\text{pos}}(I_{in}, P_{\text{pos}}) \rightarrow I_{gt}, & \text{if } M = 2, \\
\text{DMDiff}_{\text{neu}}(I_{in}, P_{\text{neu}}) \rightarrow  G(I_{gt}), & \text{if } M = 3,
\end{cases}
\end{equation}
where $P_{\text{neg}}, P_{\text{pos}}, P_{\text{neu}}$ denote the negative, positive, and neutral prompts, respectively.  
$I_{in}$ represents the input image, which can be a MetaCamera capture or a generated pseudo input image, while $I_{gt}$ is the high-quality groundtruth image. The function $G(\cdot)$ applies the edge-preserving low-pass filter.  
The variable $M$ follows a categorical distribution: 
\begin{equation}
    M \sim \text{Cat}(p_1, p_2, p_3), \quad P(M = k) = p_k.
\end{equation}
At each training step, one of the three paths, negative, positive, or neutral, is randomly selected based on the probabilities $(p_1, p_2, p_3)$.  

\subsection{Instantly Tunable Decoding}
As shown in Fig.~\ref{fig:training_and_inference}(b), in the inference step, we forward the positive and neutral-prompt paths to obtain two latent coded image \(z_{neu}\) and \(z_{pos}\). While in the decoding step, the two features are combined using a tunable scale and decoded by the VAE decoder to output the restored high-quality image:
\begin{equation}
I^* = D \left( \alpha \cdot z_{pos} + (1 - \alpha) \cdot z_{neu} \right)
\end{equation}
where \( D \) represents the VAE decoder, \( z_{pos} \) and \( z_{neu}\) denote the positive and neutral latent coded images, respectively. The parameter \( \alpha \) is a weighting factor, which controls the balance between subjective perception and objective accuracy.
In practical applications, the neural-coded images are stored as the final imaging results. When pixel-level reconstruction is needed, a series of images with varying diffusion intensities can be generated very quickly.

\section{MetaCamera Design and Implementation}

\begin{figure}[ht]
    \vspace{-10pt}
    \centering
    \includegraphics[width=0.48\textwidth]{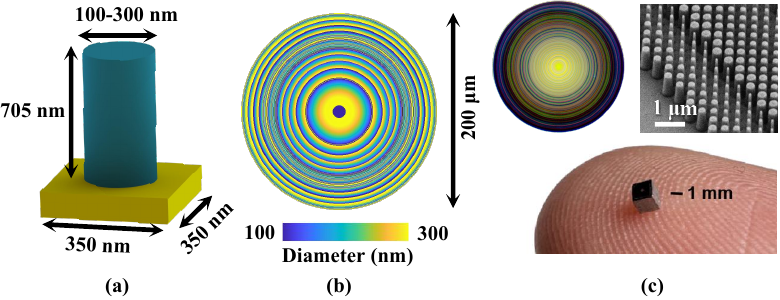}
    \caption{(a) Schematic of the metalens unit cell, consisting of a Si$_3$N$_4$ nano-pillar on a SiO$_2$ substrate. The pillar diameter varies radially to modulate the optical phase. (b) Diameter distribution map of the metalens. (c) Top: Optical and SEM images of the fabricated metalens. Bottom: Optical image of the fully integrated MetaCamera, demonstrating its ultra-compact size.}
    \label{fig:MetaCamera}
    \vspace{-10pt}
\end{figure}

Our MetaCamera integrates a metalens with a miniaturized CMOS image sensor (OV6946, OmniVision Technologies) featuring a resolution of \(400\times400\) pixels. The entire imaging module has an overall footprint of approximately $1\times1\times1$ mm$^3$, enabling ultra-compact imaging.
The metalens is composed of a dense array of subwavelength-scale nano-pillars, where the local phase modulation is dictated by the varying diameters of the cylindrical structures. 
To optimize the metalens design, we employ a combination of Finite-Difference Time-Domain (FDTD) simulations and a Neural Nano-Optics approach \cite{tseng2021neural}, ensuring precise phase control across the aperture. More details are provided in the supplementary material.

\begin{figure*}[ht]
    \centering
    \includegraphics[width=0.95\textwidth]{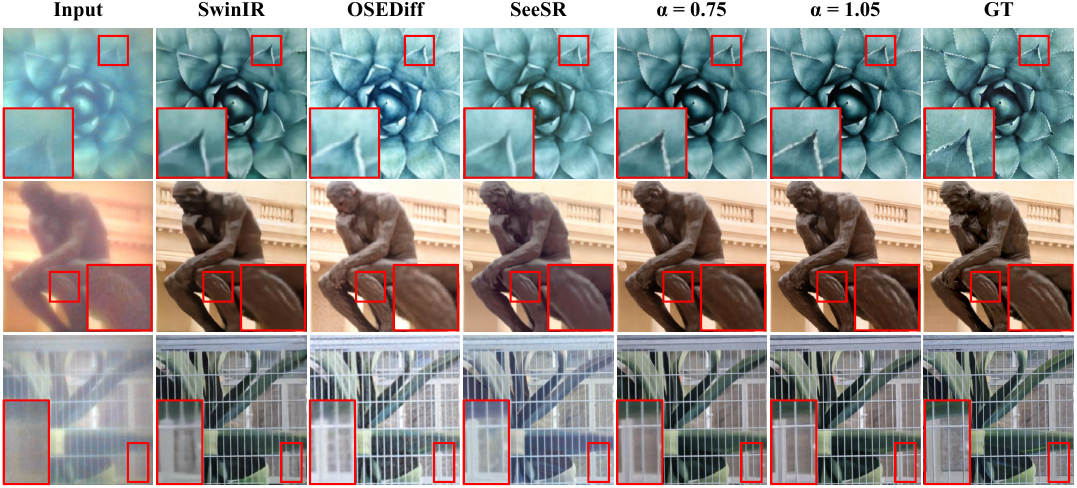} % 设置图片宽度为页面宽度
    \caption{Qualitative comparisons of different methods on our \textbf{unseen} test dataset, zoom in for details. Our method achieves high imaging quality with sharp details and maintains robust performance even in severely degraded edge regions.}
    \label{fig:exp1}
    \vspace{-10pt}
\end{figure*}
\begin{figure*}[ht]
    \centering
    \includegraphics[width=0.95\textwidth]{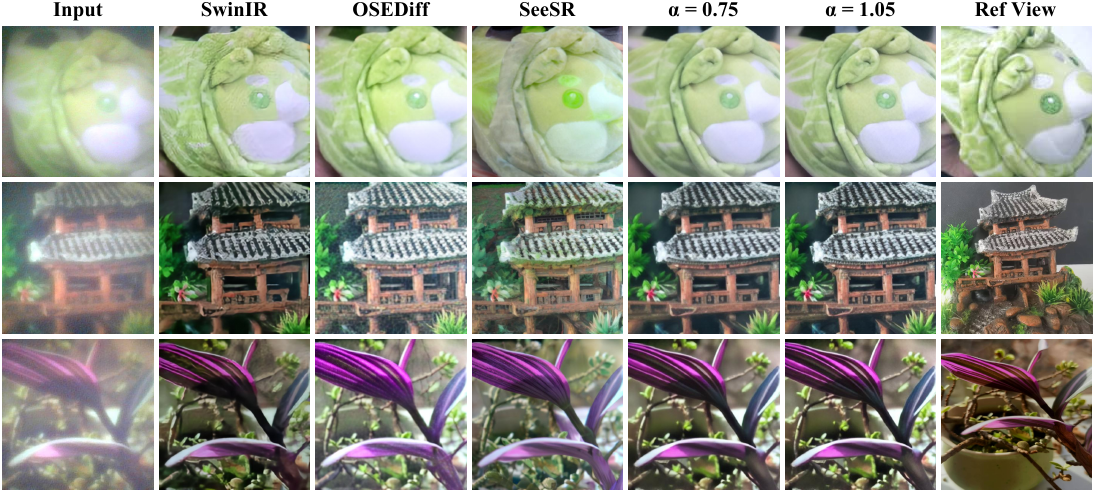} % 设置图片宽度为页面宽度
    \caption{Qualitative comparisons of different methods on real-world images captured by our system. Reference views captured by a smartphone from a similar perspective are provided. Note that smartphone software processing may introduce slight color discrepancies. Real-world images present greater challenges, highlighting the strong robustness of our method.}
    \label{fig:exp2}
    \vspace{-15pt}
\end{figure*}
\begin{figure}[ht]
    \centering
    \includegraphics[width=0.38\textwidth]{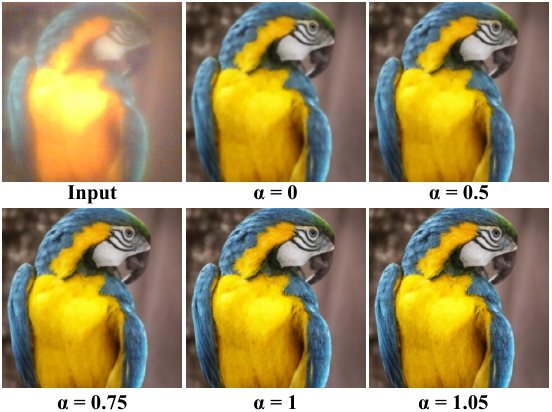} % 设置图片宽度为页面宽度
    \caption{Instantly tunable decoding demonstration. Users can dynamically adjust the diffusion intensity. }
    \label{fig:tune}
    \vspace{-10pt}
\end{figure}
\begin{figure}[ht]
    \centering
    \includegraphics[width=0.4\textwidth]{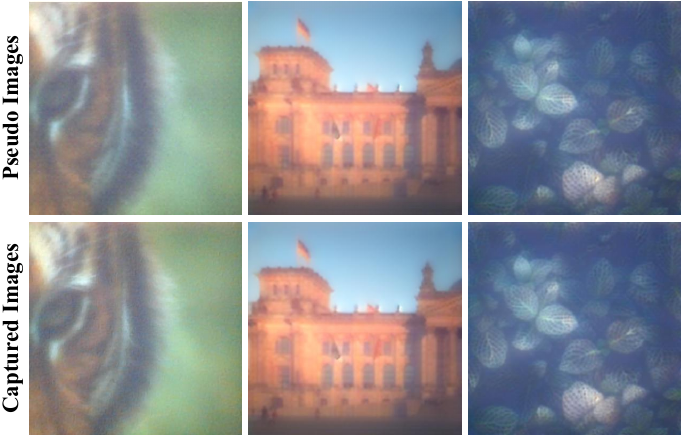} % 设置图片宽度为页面宽度
    \caption{Degradation learning qualitative results. Our method effectively simulates the imaging effects of MetaCamera. }
    \label{fig:deg}
    \vspace{-10pt}
\end{figure}
\begin{figure}[ht]
    \centering
    \includegraphics[width=0.4\textwidth]{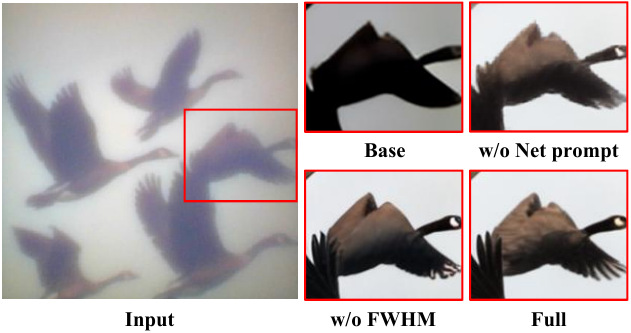} % 设置图片宽度为页面宽度
    \caption{Qualitative results of the ablation study for different modules in our method. }
    \label{fig:abl}
    \vspace{-8pt}
\end{figure}
\section{Experiment}
\begin{table*}[h]
\vspace{-4pt}
\centering
\footnotesize
\renewcommand{\arraystretch}{1.1}
\setlength{\tabcolsep}{4pt}
\begin{tabular}{l c c c c c c c c c c}
\toprule
\textbf{Methods} & \textbf{PSNR}↑ & \textbf{SSIM}↑ & \textbf{LPIPS}↓ & \textbf{DISTS}↓  & \textbf{NIQE}↓ & \textbf{MANIQA}↑ & \textbf{MUSIQ}↑ & \textbf{CLIP-IQA}↑ \\
\midrule
Wiener deconvolution\cite{wiener1964extrapolation}& 16.06 & 0.5727 & 0.6706 & 0.4393 & 20.9859 & 0.0931 & 17.41 & 0.2681 \\
Two-step PSF correction\cite{eboli2022fast}& 16.68 & 0.6251 & 0.6266 & 0.3651 &8.1287 & 0.2018 & 20.73 & 0.2851 \\
Neural nano-optics\cite{tseng2021neural}& 29.25 & 0.8624 & 0.2001 & 0.1765 &6.8027 & 0.1901 & 37.26 & 0.2746 \\
SwinIR\cite{liang2021swinir}& 29.46 & \cellcolor{red!50}\textbf{0.8786} & 0.2462 & 0.2111 & 8.2019 & 0.2208 & 36.86 & 0.3046 \\

SeeSR-s50\cite{wu2024seesr} & 23.95 & 0.8340 & 0.2315 & 0.1673 & \cellcolor{red!20}{6.6028} & 0.2633 & \cellcolor{red!20}{44.87} & \cellcolor{red!20}{0.3913} \\
OSEDiff-s1\cite{wu2025one} & 19.69 & 0.8224 & 0.2643 & 0.1868  & 7.3474 & 0.2256 & 34.52 & 0.3761 \\
\midrule
Ours-s1-{$\alpha$}0.75 & \cellcolor{red!50}\textbf{30.31}  & \cellcolor{red!20}{0.8731} & \cellcolor{red!20}{0.1705} & \cellcolor{red!20}{0.1499}  & 7.2751 & \cellcolor{red!20}{0.2782} & 44.48 & 0.3869 \\

% Ours-1-100\% & \cellcolor{red!20}{29.89} & 0.8633 & \cellcolor{red!50}\textbf{0.1504} & \cellcolor{red!50}\textbf{0.1368}  & \cellcolor{red!50}\textbf{6.4544} & \cellcolor{red!50}\textbf{0.3078} & \cellcolor{red!50}\textbf{50.63} & \cellcolor{red!50}\textbf{0.4355} \\
Ours-s1-{$\alpha$}1.05 & \cellcolor{red!20}{29.75} & 0.8598 & \cellcolor{red!50}\textbf{0.1485} & \cellcolor{red!50}\textbf{0.1356}  & \cellcolor{red!50}\textbf{6.3073} & \cellcolor{red!50}\textbf{0.3138} & \cellcolor{red!50}\textbf{51.85} & \cellcolor{red!50}\textbf{0.4460} \\

\bottomrule
\end{tabular}
\caption{Performance comparison of different methods on the dataset. Red cells indicate the best performance, and light red cells indicate the second-best performance for each metric. s1 refers to single-step diffusion, while s50 denotes diffusion with 50 steps.}
\label{tab:comparison}
\vspace{-15pt}
\end{table*}
\subsection{Experimental Setup and Dataset}
To obtain paired training data, we displayed images on a screen and captured them using our MetaCamera. Initial alignment was achieved through mechanical adjustments, followed by precise pixel-level correction via homography transformation. We also measured the MetaCamera's PSFs using an objective lens (details in the supplementary material).

To augment the dataset, we incorporated three datasets (DIV2K \cite{8014883}, Flickr2K \cite{timofte2017ntire}, and WED) with multi-scale cropping, resulting in 7,800 training images and 3,000 test images. Additionally, we generated 10,000 pseudo images via the negative diffusion process to enhance training. Beyond these datasets, we also captured real-world scene images for qualitative evaluation.

\subsection{Implementation details}

During training, we employed $L_2$ Loss and LPIPS Loss as reconstruction losses, which can be expressed as 
\begin{equation}
L = L_2 + \lambda \cdot L_{\text{LPIPS}},
\end{equation}
where the hyper-parameter $\lambda$ is set to 2.5. And the patch number $n$ in SVDA is set to 7.
The model was trained on four NVIDIA A100 80G GPUs for two days with a batch size of 16. For a fair comparison, all baseline methods were trained using the same dataset. More details are delivered in the supplementary material.

\subsection{Image quality Assessment}

We compare our method with several baselines, including PSF-based non-blind denoising (Wiener deconvolution, neural nano-optics\cite{tseng2021neural}), kernel-based blind denoising (Two-step PSF correction\cite{eboli2022fast}), transformer-based restoration (SwinIR~\cite{liang2021swinir}), and diffusion-based methods. Specifically, we evaluate SeeSR, which leverages ControlNet for structure-aware restoration, and OSEDiff, a single-step denoising model based on LoRA fine-tuning. All methods are trained on the same dataset.
For evaluation, we use both reference and non-reference metrics. PSNR and SSIM~\cite{wang2004image} measure fidelity, while LPIPS~\cite{Zhang2018b}, and DISTS~\cite{Ding2020} assess perceptual quality. We also include NIQE~\cite{6353522}, MANIQA~\cite{yang2022maniqa}, MUSIQ~\cite{Ke2021}, and CLIP-IQA~\cite{wang2022exploring} for non-reference evaluation.

Table \ref{tab:comparison} shows that our method surpasses all baselines across all metrics, demonstrating its effectiveness. The blind denoising method and simple Wiener deconvolution struggle to restore the image due to sensor noise and chromatic aberration. We also performed two qualitative comparisons, with one using dataset images, as shown in Fig.~\ref{fig:exp1}, and the other using real-world images, as shown in Fig.~\ref{fig:exp2}. From the qualitative and quantitative results, 
it can be observed that non-diffusion-based methods, such as SwinIR, produce blurry images lacking high-frequency details due to the absence of priors. However, this also preserves accurate color tones and low-frequency structures, leading to higher PSNR and SSIM but poorer subjective perception scores. In contrast, diffusion-based methods generate richer high-frequency details and more natural images by leveraging priors. Yet, their reliance on priors and lack of degradation modeling introduces errors in color tones, low-frequency structures, and sometimes incorrect details.

Our method combines the strengths of both, achieving accurate color tones, rich details, and enhanced realism, while the instantly tunable decoding strategy enables users to adjust the diffusion style as needed dynamically. Notably, due to its robust modeling of optical characteristics, our method maintains high performance even at the edges of images, where other methods typically degrade significantly. Furthermore, the strong chromatic dispersion of metasurfaces poses challenges for diffusion models incorporating text semantics, often leading to instability. In contrast, our approach relies solely on imaging quality descriptions in the prompt, avoiding dependence on scene content and ensuring consistent performance.

The real-world dataset used in Fig.~\ref{fig:exp2} presents more complex lighting conditions and noise characteristics. Despite these challenges, our method exhibits remarkable robustness, whereas other approaches struggle to maintain effectiveness in practical applications.

\begin{table}[ht]
\vspace{-5pt}
\centering
\footnotesize
\renewcommand{\arraystretch}{1.1}
\setlength{\tabcolsep}{1.8pt}
\begin{tabular}{l c c c c c c c c c c}
\toprule
\textbf{Methods} & \textbf{PSNR}↑ & \textbf{SSIM}↑ & \textbf{LPIPS}↓ & \textbf{MANIQA}↑ & \textbf{MUSIQ}↑ \\
\midrule
Ours-{$\alpha$}0 & 30.06  & 0.8667 & 0.2715  &0.2276 & 31.24 \\

Ours-{$\alpha$}0.5 & \cellcolor{red!50}\textbf{30.39}  & \cellcolor{red!50}\textbf{0.8743} & 0.2039  &0.2557 & 38.96 \\
Ours-{$\alpha$}0.7 & 30.31  & 0.8731 & 0.1705  & 0.2782 & 44.48 \\
Ours-{$\alpha$}0.9 & 30.10  & 0.8685 & 0.1562  &0.2956 & 48.14 \\

Ours-{$\alpha$}1 &29.89 & 0.8633 &0.1504 & 0.3078  & 50.63 &  \\
Ours-{$\alpha$}1.05 & 29.75 & 0.8598 & \cellcolor{red!50}\textbf{0.1485} &  \cellcolor{red!50}\textbf{0.3138} & \cellcolor{red!50}\textbf{51.85} &  \\

\bottomrule
\end{tabular}
\caption{Performance comparison of different diffusion injection factors.}
\label{tab:ab1}
\vspace{-12pt}
\end{table}

\begin{table}[ht]

\centering
\footnotesize
\renewcommand{\arraystretch}{1.1}
\setlength{\tabcolsep}{1.8pt}
\begin{tabular}{l c c c c c c c c c c}
\toprule
\textbf{Methods} & \textbf{PSNR}↑ & \textbf{SSIM}↑ & \textbf{LPIPS}↓ & \textbf{MANIQA}↑ & \textbf{MUSIQ}↑ \\
\midrule
Base & 17.12  & 0.7685 & 0.3455  &0.2332 & 38.27 \\
w/o FWHM&  26.62  & 0.8414& 0.1869  &0.2966& 50.55\\
w/o Neg prompt&  28.21  & 0.8571 & 0.1953  &0.2587 & 44.15\\
Ours-{$\alpha$}1 &\cellcolor{red!50}\textbf{29.89} & \cellcolor{red!50}\textbf{0.8633} &\cellcolor{red!50}\textbf{0.1504} & \cellcolor{red!50}\textbf{0.3078}  & \cellcolor{red!50}\textbf{50.63} &  \\
\bottomrule
\end{tabular}
\caption{Quantitative results of the ablation study for different modules in our method.}
\label{tab:ab2}
\vspace{-18pt}
\end{table}

\subsection{Ablation Study}

The results in Table \ref{tab:ab1} and Fig.~\ref{fig:tune} validate the effectiveness of our multipath diffusion and instantly tunable decoding strategies. Experiments with various injection ratios reveal that higher diffusion injection strengths enhance subjective perception metrics such as LPIPS, MANIQA, and MUSIQ, producing images that are more vivid and detailed, albeit with some excessive detail, whereas lower injection strengths favor the generation of conservative low-frequency structures, thereby improving PSNR and SSIM. 
% Notably, the instantly tunable decoding mechanism allows for the adjustment of the image generation style according to user preferences.

To evaluate the effectiveness of various components in our method, we performed an ablation study by selectively removing key modules and observing the impact on performance. Table \ref{tab:ab2} and Fig.~\ref{fig:abl} presents the results.
Without any of our proposed modules, a simple LoRA fine-tuning base model fails to achieve effective image restoration. Moreover, removing any individual module leads to a decline in both quantitative metrics and visual quality, further validating the effectiveness of our method.
 
% To evaluate the effectiveness of various components in our method, we performed an ablation study by selectively removing key modules and observing the impact on performance. Table \ref{tab:ab2} presents the results, which assess the contributions of SVDM attention and negative prompt training. The removal of either module leads to a decline in numerical performance, as indicated by increased artifacts and reduced fidelity; their joint removal produces the worst results, underscoring their complementary roles in enhancing spatial adaptiveness and overall image quality.

Furthermore, Fig.~\ref{fig:deg} qualitatively demonstrates the effectiveness of degradation learning, indicating that our method accurately simulates the imaging effects of MetaCamera.

\vspace{-0.2cm}
\section{Conclusion}

We proposed a diffusion-based framework for high-quality metalens photography. Our approach introduced an SVDA module for optical and sensor degradation awareness. A multipath diffusion strategy and an instantly tunable decoder further enhanced reconstruction quality and adaptability. Experiments showed that our method outperformed existing approaches, setting a new standard for high-quality metalens photography and advancing computational imaging.

% We propose a diffusion-based framework for high-quality metalens photography. Our approach introduces an SVDA module for optical and sensor degradation awareness. A multipath diffusion strategy and an instantly tunable decoder further enhance reconstruction quality and adaptability. Experiments show that our method outperforms existing approaches, setting a new standard for high-quality metalens photography and advancing computational imaging.

% In conclusion, we introduce a computational imaging system based on Diffusion-Powered Nanooptics. Our approach pioneers the first metasurface-based miniature imaging system by integrating an ultra-thin metasurface lens into a compact camera. By fine-tuning a Stable Diffusion model with LoRA and employing a Spatially Varying Degradation Modeled Attention module, we achieve high-quality, single-step image restoration that compensates for spatial distortions. Additionally, our Multipath Diffusion Training and tunable decoding strategy effectively balance perceptual quality and reconstruction accuracy, setting a new benchmark for ultra-compact imaging systems.

{
    \small
    \bibliographystyle{ieeenat_fullname}
    \bibliography{main}
}


\end{document}